%% file: main.tex
\documentclass[10pt,twocolumn,letterpaper]{article}

\usepackage{iccv}
\usepackage{times}
\usepackage{epsfig}
\usepackage{graphicx}
\usepackage{amsmath}
\usepackage{amssymb}
\usepackage[T1]{fontenc}
\usepackage{mathrsfs}
\usepackage{booktabs}
\usepackage{multirow}
\usepackage{comment}
\usepackage{array}
\usepackage{colortbl}
\usepackage{booktabs}
\usepackage{enumerate}
\newcommand{\eat}[1]{}

\iccvfinalcopy 

\def\iccvPaperID{****} 

\ificcvfinal\fi
\begin{document}

\title{Frame-Recurrent Video Inpainting by Robust Optical Flow Inference}


\author{
Yifan Ding$^1$\quad
Chuan Wang$^2$\quad 
Haibin Huang$^2$\quad
Jiaming Liu$^2$\quad
Jue Wang$^2$\quad
Liqiang Wang$^1$\\
{\normalsize$^1$University of Central Florida}\\
{\tt \normalsize yf.ding@knights.ucf.edu, lwang@cs.ucf.edu}\\
{\normalsize $^2$Megvii Technology}\\
{\tt \normalsize \{wangchuan, huanghaibin, liujiaming, wangjue\}@megvii.com}
}

\maketitle

\input{abs.tex}

\input{intro.tex}

\input{related_work.tex}

\input{algo.tex}

\input{results.tex}

\input{conclu.tex}
{\small
\bibliographystyle{ieee}
\bibliography{main}
}

\end{document}

%% file: abs.tex
\begin{abstract}\label{sec:abs}

In this paper, we present a new inpainting framework for recovering missing regions of video frames. Compared with image inpainting, performing this task on video presents new challenges such as how to preserving temporal consistency and spatial details, as well as how to handle arbitrary input video size and length fast and efficiently. Towards this end, we propose a novel deep learning architecture which incorporates ConvLSTM and optical flow for modeling the spatial-temporal consistency in videos. It also saves much computational resource such that our method can handle videos with larger frame size and arbitrary length streamingly in real-time. Furthermore, to generate an accurate optical flow from corrupted frames, we  propose a robust flow generation module, where two sources of flows are fed and a flow blending network is trained to fuse them. We conduct extensive experiments to evaluate our method in various scenarios and different datasets, both qualitatively and quantitatively. The experimental results demonstrate the superior of our method compared with the state-of-the-art inpainting approaches.

\end{abstract}

%% file: intro.tex
\begin{figure}[t]
  \centering
  \includegraphics[width=0.98\linewidth]{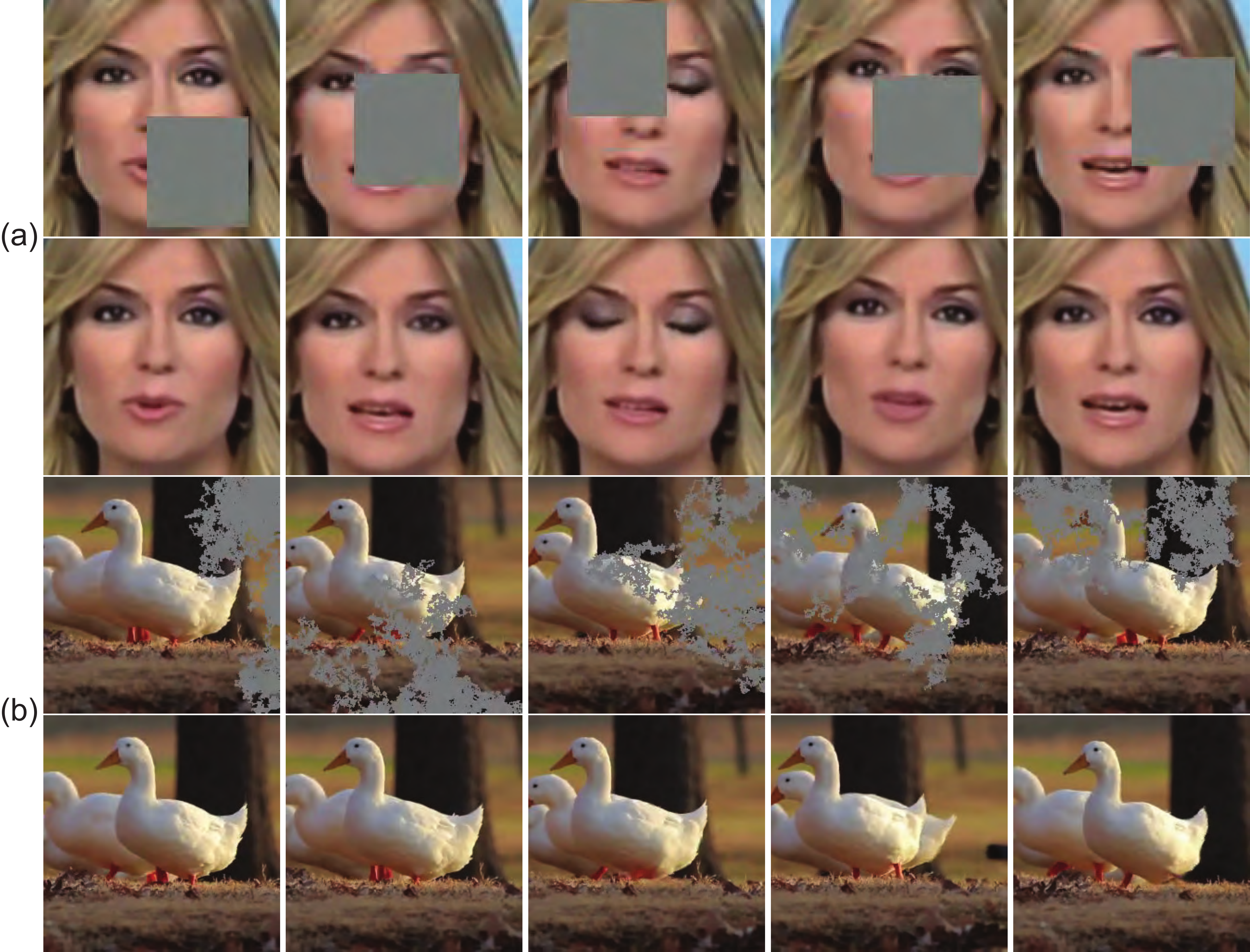}\\
  \vspace{5pt}
  \caption{Video inpainting results by our method. Row 1 and 3 are consecutive input frames with random rectangle and random walker masks. Row 2 and 4 are our results which contain rich visual details and are temporally consistent.}\label{fig:teaser}
\end{figure}

%

\section{Introduction}\label{sec:intro}

Video creating and editing are becoming increasingly popular nowadays due to rapid development of Internet. However, it remains a big challenge for the task so called video completion, i.e. inpainting, which aims at filling missing regions caused by unwanted object removal or corruption of video data in transmission, and recovering plausible frames. Due to the additional temporal dimension, directly applying existing image inpainting techniques frame by frame is commonly problematic because it lacks the preservation of inter-frame consistency of the input video and suffers from flickering artifacts .


With the development of deep neural network in video tasks~\cite{wang2019gif2video,meng2018mganet,nilsson2018semantic}, early trial of extending 2D image inpainting methods into video inpainting via 3DCNNs is proposed in~\cite{wang2018video}. In that work, a 3DCNN module is introduced to recover the inter-frame temporal coherence and used as a guidance for a 2DCNN module which generates high-quality frames and outputs recovered videos. However, due to the memory limits and high computational cost of 3DCNN, that approach can deal with videos of fixed-size only and a streaming adaption of 3DCNN is far from straight-forward. 3DCNN also suffers from its limited capability of recovering motions in video because the kernel size in 3DCNN limits the motion range it models. These drawbacks disable it being further applied to a variety of video data.

In this work, we tackle the problem of video inpainting by solving the following issues: preserving temporal consistency and spatial details as well as handling arbitrary input video size and length efficiently. Inspired by~\cite{lai2018learning,liu2016spatio,gao2017video}, we propose a ConvLSTM based video inpainting method which uses image based algorithms to model spatial information intra-frame and recurrent neural networks to model temporal information inter frame with optical flow as an intermediary.  Incorporating these two modules, our model circumvents the memory problems brought by 3DCNN and can handle videos with larger motion thanks to the optical flow which is not constrained by the kernel size.

A problem left to be solved in our model is that optical flow is hard to gain region in holes inside the frames. In our model, 
we incorporate a robust optical flow generation module to obtain a trustable optical flow to guide  the output of an image inpainting algorithm the ConvLSTM. In this module, two sources of initially recovered optical flows, one by the inpainted frames and the other by the inpainted optical flow, are fused together by a blending network. The output flow commonly has higher accuracy which enables larger motion prediction. The ConvLSTM based network further models the temporal information within the inpainted frames and gradually remove the inter-frame flickers through a loss considering the temporal and spatial balance.


To verify the effectiveness of the network, we tested our model on two datasets, the FaceForensics~\cite{rossler2018faceforensics} of faces with moving emotion expressions and the DAVIS+VIDEVO~\cite{lai2018learning} being made of moving objects on natural backgrounds. The testing involves three types of masks, fixed rectangles, random rectangles or a more complicated random walker masks. Experimental results reveal the superior of our method than state-of-the-art. Ablation studies also demonstrate the ability of the ConvLSTM module to model temporal information as well as the robustness of the optical flow generation module.

In summary, our main contributions include:
\begin{itemize}
\setlength\itemsep{-2pt}
\item A ConvLSTM based video inpainting network which combines the temporal and spatial constraints to reconstruct videos with details but without flickering artifacts.
\item A robust motion estimation network via trainable optical flow fusion module which enables the prediction of large motions so as to play a key role in the success of our method. 
\item State-of-the-art performance in the video inpainting task, both in quality and efficiency under three types of masks. Our method is able to handle videos of arbitrary length and frame size, which is a limitation of previous work.
\end{itemize}

%% file: related_work.tex
\section{Related Work}\label{sec:related_work}
\input{pipeline.tex}

\paragraph{Patch-based image/video inpainting.}
Patch-based synthesis is the most used traditional strategy for image inpainting. It was firstly proposed in~\cite{efros1999texture} and later improved by works~\cite{kwatra2005texture,wexler2007space,barnes2009patchmatch,barnes2015patchtable}. The basic idea is to recover the missing contents in a region-growing way, i.e. the algorithms start from boundary of holes and extend the region by searching appropriate patches and assembling them together. The improved methods work on different directions in searching and optimization, or for application like face~\cite{zhao2018identity,yamaguchi2018high}.
It is also adapted to video inpainting problem by replacing 2D patch synthesis with 3D spatial-temporal patch synthesis across frames. This
was firstly proposed in~\cite{wexler2004space,wexler2007space} to ensure the temporal consistency of the generated video and later improved in~\cite{jia2005video,venkatesh2009efficient} to handle more complicated video input.  However, all of these works are designed for the video with repeated content across frames. They are unable to tackle the problem we proposed in this paper where missing parts cannot be replaced by similar content in the input. Resorting to a large video dataset, we try to train a ConvLSTM in this work for missing contents prediction based on the high-level spatial-temporal context understanding. 

\paragraph{CNN-based image/video inpainting.}
Recently, Convolutional Neural Network was applied to image inpainting for small holes~\cite{xie2012image} and then extended to larger missiong regions~\cite{pathak2016context}. Later on, Yang et al.~\cite{yang2017high} further developed a multi-scale neural patch synthesis algorithm that not only preserves contextual structure but also produces high-frequency details. The algorithm proposed in~\cite{IizukaSIGGRAPH2017} further enhances the performance by involving two adversarial losses to measure both the global and local consistency of the result. Different from the previous works which only focus on box-shaped holes, This method also develops a strategy to handle the holes with arbitrary shapes. Liu et al. further solve the inpainting problem by introducing a partial convolution operator~\cite{liu2018image}, to better handle challenging holes generated by a random walker.

Wang et al. extended the CNN solution from image to video, proposing a hybrid network combining 2DCNN and 3DCNN. They used 3DCNN to recover the temporal coherence and 2DCNN to reconstruct the spatial details. However, due to the heavy computational cost and limited power to learn temporal information by 3DCNN, this method works only for videos with limited frame size and length. It cannot handle videos with medium motion either. On the contrary, our method removes the restriction of video length and frame size, and is able to better model the temporal consistency by a ConvLSTM module and a robust flow generation network.

\paragraph{Learning temporal consistency by LSTM.}
It has been well demonstrated in recent years that recurrent neural network (RNN), or its variants Long Short-Term Memory (LSTM)~\cite{hochreiter1997long,gers1999learning,ren2016look} and ConvLSTM~\cite{xingjian2015convolutional} are powerful tools to model long-term temporal consistency for video related tasks, e.g. video captioning~\cite{gao2017video} and action recognition~\cite{liu2016spatio}. The power of ConvLSTM results from its self-parameterized controlling gates which can decide a memory cell "remember" or "forget" the present and past information, which is commonly formulated as the following equations:
\begin{align*}
        i_t &= \sigma(W_{xi} \circledast X_t + W_{hi} \circledast H_{t-1} + W_{ci} \circ C_{t-1} + b_i)\\
        f_t &= \sigma(W_{xf} \circledast X_t + W_{hf} \circledast H_{t-1} + W_{cf} \circ C_{t-1} + b_f)\\
        c_t &= f_t \circ C_{t-1} + i_t \circ \tanh(W_{xc} \circledast X_t + W_{hc} \circledast H_{t-1} + b_c)\\
        o_t &= \sigma(W_{xo} \circledast X_t + W_{ho} \circledast H_{t-1} + W_{co} \circ C_{t}  + b_o) \\
        H_t &= o_t \circ \tanh(C_t)
\end{align*}
where $\circ$ is the Hadamard product and $\circledast$ is the convolution operator.

In~\cite{xingjian2015convolutional}, the capacity of ConvLSTM is well demonstrated by the task of weather forecast, where spatial-temporal information in satellite images is modeled by the combination of convolution layers and LSTM. Additionally, optical flow conveys motion information which can propagate the results in previous frame into the current one. It usually serves as an assistant to the ConvLSTM based tasks, e.g. video segmentation~\cite{nilsson2018semantic, terwilliger2019recurrent}, action recognition~\cite{das2018deep,li2018videolstm} and detection~\cite{zhou2017spatio}. Following the similar idea, in this paper we also apply ConvLSTM and optical flow to video inpainting task, to demonstrate their effectiveness.

%% file: pipeline.tex
\begin{figure*}[t]
  \centering
  \includegraphics[width=0.98\linewidth]{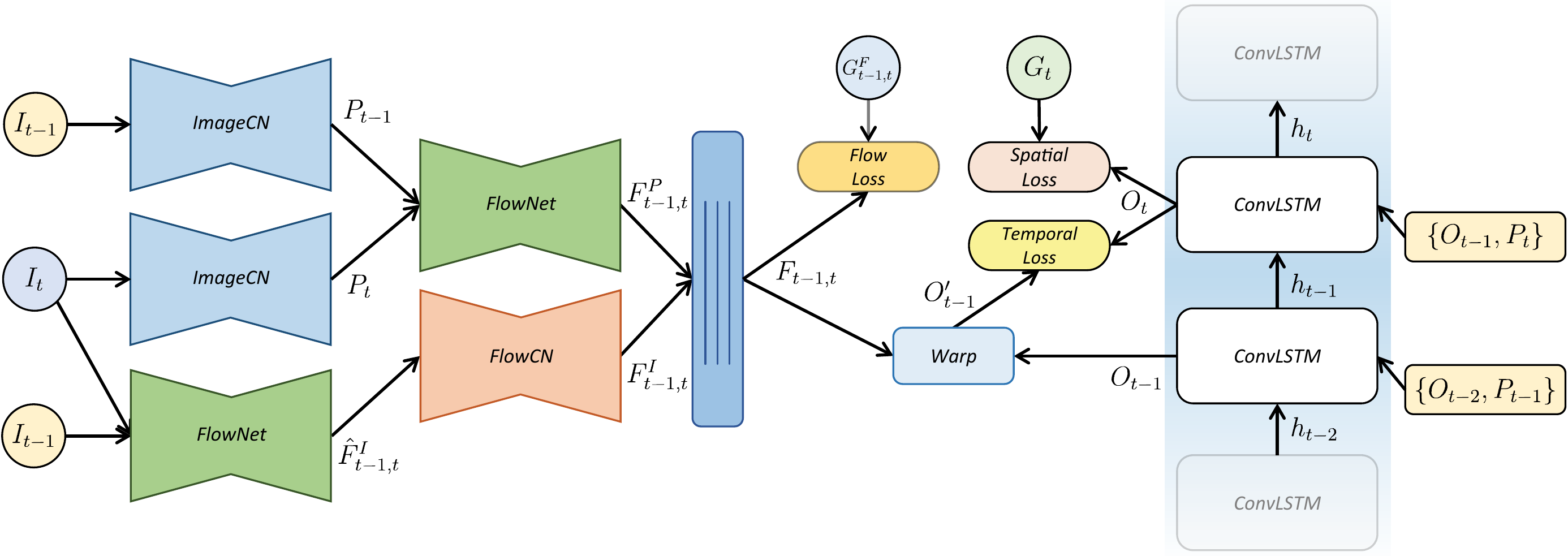}\\
  \vspace{6pt}
  \caption{Network structure. $\{I_t\}$ denotes the input frames with holes inside. $\{P_t\}$ denotes the inpainted frames using image inpainting method. $\{F^P\}$ and $\{F^I\}$ are the two branch optical flow generated from $\{P_t\}$ and $\{I_t\}$ separately. $\{F_t\}$ is the final output of the Blending network which is illustrated in Figure~\ref{fig:blending-network}}\label{fig:pipeline}
\end{figure*}

%% file: algo.tex
\section{Algorithm}\label{sec:algo}

\paragraph{Overview.}
Our method is built upon deep neural networks, including CNN and ConvLSTM. It streamingly takes two successive frames $\{I_{t-1}, I_t\}$ of an incomplete video $V_{in}$ as input, and produces a complete frame $O_t$ as output, where $O_t$ contains spatial details and holds temporal consistency with last time output, $O_{t-1}$.

To achieve the two goals simultaneously, we design a framework composed of three functional parts. The first part is an image-based inpainting module $\mathcal H_s$, produces an inpainted frame $P_t$ given $I_t$, i.e.
\begin{equation*}\label{eq:Gs}
  P_t = \mathcal H_s(I_t)
\end{equation*}
which is achieved by an existing image inpainting algorithms such as Partial Convolution~\cite{liu2018image} which is used in this paper, where the produced frames $\{P_t\}$ normally contain flickering artifacts. Then we recurrently feed a pair of results $\{O_{t-1}, P_t\}$ to a ConvLSTM module $\mathcal H_t$, to produce the output $O_t$ of the current time stamp $t$, which is introduced in details in Section~\ref{sec:convlstm}. To ensure $O_t$ to be visually plausible to its ground truth $G_t$ and coherent with $O_{t-1}$, we involve newly designed losses and adopt several training strategies (Section~\ref{sec:loss}), which rely on a robust optical flow generation module $\mathcal H_f$ as introduced in Section~\ref{sec:flow-generation}. To give an overview, we illustrate the entire pipeline of our framework in~\figurename~\ref{fig:pipeline}. 

%
%
%
%
%
%

\subsection{Robust Flow Generation}\label{sec:flow-generation}

Optical flow plays a key role in video related tasks as it provides direct access to the temporal information. However, with holes in the frames, estimating the optical flow becomes far from straight-forward for video inpainting task. To tackle this issue, we consider two separated paths to generate a robust optical flow which we can rely on for the output temporal consistency.

We manage to consider two sources of the optical flow. One branch for the optical flow, i.e. $F_{t-1,t}^P$, is generated from the separately inpainted frames $\{P_t\}$ by an optical flow estimation module $\mathcal{F}_l$, i.e.
\begin{equation*}
F_{t-1,t}^P = \mathcal{F}_l(P_{t-1}, P_t)
\end{equation*}
The other branch, i.e. $F_{t-1,t}^I$, results from the completion of defective optical flow $\hat{F}_{t-1,t}^I$ generated from the input frames $\{I_t\}$, by an optical flow completion network $\mathcal{H}_c$, i.e.
\begin{align*}
F_{t-1,t}^I = \mathcal{H}_c(\hat{F}_{t-1,t}^I), \qquad \hat{F}_{t-1,t}^I = \mathcal{F}_l(I_{t-1}, I_t)
\end{align*}

Coming from different paths, $F_{t-1,t}^I$ and $F_{t-1,t}^P$ normally have various distributions. Specifically, $F_{t-1,t}^P$ looks smoother but sometimes brings in artifacts that may mislead ConvLSTM to correct the inpainting results. On the contrary, $F_{t-1,t}^I$ accords to the statistical characteristics of the optical flow in the training dataset, but it occasionally fails to be well filled and a border is usually shown in the inpainted optical flow. To take advantage of the two flows, we design a Flow Blending Network $\mathcal{H}_f$ to generate a more robust flow.


\input{blend-network.tex}
\paragraph{Flow blending network $\mathcal{H}_f$.} We propose Flow Blending Network $\mathcal{H}_f$ to blend the two optical flows so as to remove the errors. Its structure is illustrated in Figure~\ref{fig:blending-network}, where the input is a channel-wisely stacked version of $\{F^P_{t-1,t},F^I_{t-1,t}\}$ and the output is a refined flow $F_{t-1,t}$.
$\mathcal H_f$ adapts an U-Net~\cite{ronneberger2015u} architecture where the extracted features from the encoder are concatenated to the corresponding decoder layers to enhance the performance. To produce $F_{t-1,t}$, a residual value is predicted and added to $F^P_{t-1,t}$ and $F^I_{t-1,t}$, i.e.
\begin{equation}\label{eq:flow}
  F_{t-1,t} = \frac{1}{2}\left(F^P_{t-1,t} + F^I_{t-1,t} + \mathcal H_f(F^P_{t-1,t},F^I_{t-1,t})\right)
\end{equation}

Note that, there is still other options for the design of such a flow blending network. For example, the attention mechanism~\cite{xu2015show} uses fully connected layers to weight the feature maps, providing more flexibility. However, due to large number of parameters involved, this method requires more computational resources which potentially limits the efficiency for video inpainting. On the contrary, with only 6 convolutional layers (3 for the encoder and 3 for the decoder), our flow blending network significantly limits the number of parameters, making it invariant with the size of the feature maps. Together with the ConvLSTM module which is also built upon FCN, our network can accept arbitrary size of frames theoretically.

With the ground truth optical flow $G_{t-1,t}^F$ as a supervision, our flow blending network $\mathcal{H}_f$ can learn an optimized blending mechanism for the two source optical flows. This process can be written as:
\begin{align*}
\min_{\mathcal{H}_f}~L_f = \sum_{t=1}^T || G_{t-1,t}^F - F_{t-1,t} ||_1 \label{eq:loss-flow}
\end{align*}
where $L_f$ is the sum of $l_1$ loss between $G_{t-1,t}^F$ and $F_{t-1,t}$ over all $T$ time steps. We demonstrate the effectiveness of $\mathcal{H}_f$ in Table~\ref{tab:b1-b2-lstm-pc-ours} and Paragraph~\ref{par:one-flow}.

\paragraph{Optical flow completion network $\mathcal{H}_c$.} We follow a similar idea to inpaint the optical flow by designing an optical flow completion network. The modification lies in that, unlike imagery data, optical flow has different value range and channel numbers, so that special training schemes have to be designed for it. To be specific, we first normalize the optical flow data by an instance-wise value range instead of a constant range as usually adopted for RGB image data. {\color{black}{Additionally, we transform the two-channel flow data to three channels to utilize the pre-trained weights, where the third channel is filled with the mean value of the first two channels.  } }

\subsection{Recurrent Neural Network}\label{sec:convlstm}

With the single inpainted frames $\{P_t\}$ and the blended optical flow $\{F_{t-1,t}\}$ computed, we further apply Recurrent neural network~\cite{rumelhart1988learning} to enforce the temporal consistency and remove inter-frame flickers in $\{P_t\}$.

Specifically, we adopt ConvLSTM to capture the spatial-temporal information in the videos. We first pass $O_{t-1}$ and $P_t$ through several convolutional layers separately, and then concatenate their extracted features and feed them to the ConvLSTM module $\mathcal{H}_t$. Finally, the output state of the ConvLSTM is decoded to the residual value between $O_{t-1}$ and $O_t$, i.e.
\begin{equation}
O_t = O_{t-1} + \mathcal H_t(O_{t-1},P_t)
\end{equation}

With ConvLSTM involved, the temporal correlation between two successive frames are well modeled. Since it does not restrict the length of the input sequences, our method can work in a streaming manner unlike~\cite{wang2018video}. Furthermore, with fully convolutional modules inside, the spatial details can be also reconstructed to produce accurate and smooth frames.



\subsection{Training Losses and Strategy}\label{sec:loss}

\paragraph{Spatial losses.} We compute the $l_1$ distance between the output frames $\{O_t\}$ and the ground truth frames $\{G_t\}$ to guide our model spatially reconstructing the frames. Besides the per-pixel loss, we also incorporate perceptual losses~\cite{johnson2016perceptual} between $\{O_t\}$ and  $\{G_t\}$. Specially, the two losses are
\begin{align}
  L_d &= \sum_{t=1}^T{M_t\cdot{||O_{t}-G_t||_1}} \label{eq:loss-l1} \\
  L_p &= \frac{1}{N} \sum_{l=1}^N{\sum_{t=1}^T{|| \mathcal H_p(O_t) -  \mathcal H_p(G_t)  ||_1}} \label{eq:loss-percept}
\end{align} 
where $\mathcal{H}_p$ is a pre-trained feature extractor such as ResNet~\cite{he2016deep} or VGG-16~\cite{simonyan2014very} as used in this paper, and $N$ is the number of feature maps of $\mathcal{H}_p$. 



\paragraph{Short-term temporal losses.} The temporal losses are proposed to maintain the temporal consistency between consecutive output frames. We first compute the $l_1$ loss between $O_t$ and $O'_{t-1}$, a warped version of $O_{t-1}$, to enforce the transition from $O_{t-1}$ to $O_t$ as smooth as possible. Specifically, it is calculated as
\begin{align}
  L_s =  \sum_{t=2}^T{M_t\cdot{||O_{t}-O'_{t-1}||_1}} \label{eq:loss-short-term}
\end{align}
where $M_t$ is a mask representing the missing pixels in the input frame, i.e. regions with $M_t = 1$ need to be recovered. The warping operation is achieved by re-mapping the pixel values of $O_{t-1}$ by flow $F_{t-1,t}$, i.e.
\begin{equation}
O'_{t-1} = Warp(O_{t-1}, F_{t-1,t}) \label{eq:warpping}
\end{equation}
which naturally supports gradient back-propagation.

$L_s$ encodes a short-term temporal loss which involves two consecutive frames in the forward direction. In practice, we also apply the aforementioned process to the reverse version of the input video, producing another short-term temporal loss, i.e.
\begin{align}
  L_r =  \sum_{t=2}^T{M_t\cdot{||O_{t-1}-O'_{t}||_1}} \label{eq:loss-reverse}
\end{align}
With bi-directional temporal losses involved, the results could be potentially optimized in a more stable and rarely error-prone manner.


%
%
%

\paragraph{Long-term temporal losses.} We also calculate a long-term temporal loss between $O'_{t}$ and $O_1$ and $O_T$, respectively, to maintain the consistency of the whole video as well rather than just two consecutive frames. Similarly, it is written as 
\begin{align}
  L_l = \sum_{t=1}^T M_t \cdot \left( ||O_{1}-O'_{t}||_1 + ||O_{T} - O'_{t}||_1 \right) \label{eq:loss-long-term}
\end{align}

%


\paragraph{Training.} During training, we weight and sum up the aforementioned losses for a global optimization, covering temporal and spatial ones. Specifically, the overall loss of our model is defined as:
\begin{equation}\label{eq:loss-overall}
  \mathbf{L} = \sum_{i\in U}{ \lambda_i \cdot L_i } , \quad U = \{s,f,d,p,r,l\}
\end{equation}
where $\lambda_i$ denotes the weight of the corresponding loss term $L_i$, which is manually pre-defined in the experiments. In practice, to facilitate the optimization, we first pre-train the image inpainting and flow inpainting modules to a stable status, and then train the ConvLSTM and the Flow Blending Network without finetuning the two inpainting modules. With all of the training losses and strategies as proposed, our network is able to inpaint frames with plausible details but without flickering artifacts.

%% file: blend-network.tex
\begin{figure}[t]
\begin{center}
  \includegraphics[width=0.9\linewidth]{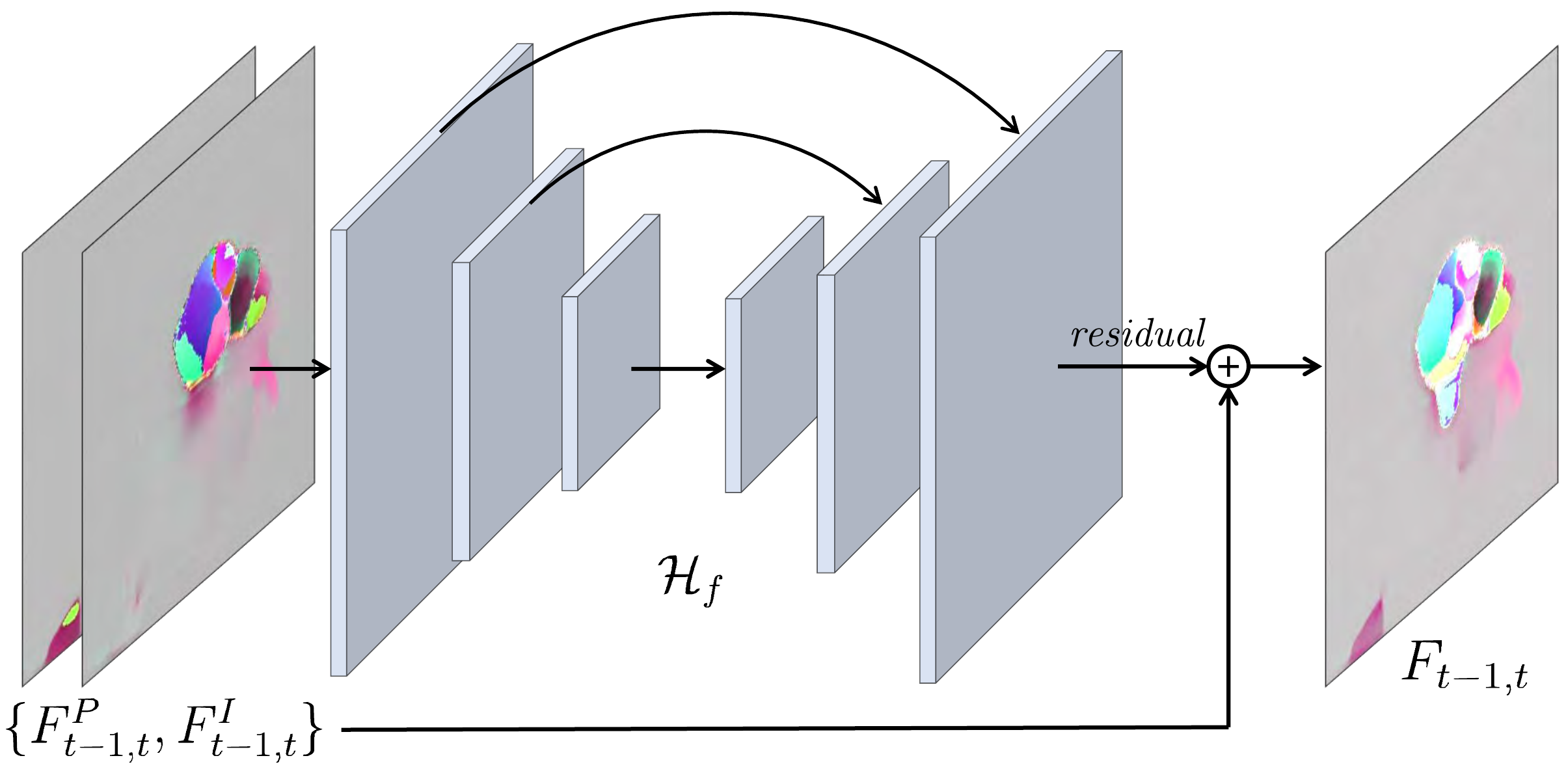}
  \end{center}
  \caption{Structure of Flow Blending Network. We use a three layers encoder-decoder architecture and the feature maps from the encoder are concatenated to the corresponding decoder layers to improve the performance. The input flow pairs are stacked channel-wisely. }\label{fig:blending-network}

\end{figure}

%% file: results.tex
\input{results_fig.tex}
\input{tab-accuracy.tex}

\input{tab-efficiency.tex}
\input{ours-vs-lstm-pc.tex}
\section{Experimental Results}\label{sec:results}
\subsection{Dataset and Experimental Settings}



We test our network on two datasets and three types of masks. FaceForensics~\cite{rossler2018faceforensics} is a human face dataset containing 1,004 video clips with near frontal pose and neutral expression changed across frames. To fully excavate the potential of our framework, we also test on the DAVIS+VIDEVO dataset~\cite{lai2018learning} which has 190 videos and contains a variety of moving objects and motion types. 
The three types of masks include: \vspace{-1mm}
\begin{enumerate}[1)]
\setlength\itemsep{-2pt}
    \item Fixed rectangles: the rectangle masks are the same across all frames in a video;
    \item Random rectangles: each frame in a video have rectangle mask of changing size and locations;
    \item Random walker: the masks have random streaks and holes of arbitrary shapes as in~\cite{sundaram2010dense}
\end{enumerate}\vspace{-1mm}
For the two rectangle masks, we follow the setting in~\cite{wang2018video} which generates masks of size between $[0.375l, 0.5l]$, where $l$ is the frame size.

In all experiments, we use FlowNet2~\cite{ilg2017flownet} for online optical flow generation. As mentioned, we also pre-train the frame inpainting network $\mathcal{H}_s$ and the flow inpainting network $\mathcal{H}_c$ to boost the training process. We use Adam~\cite{kingma2014adam} as the optimizer and set the learning rate to be $1.0\times10^{-4}$. For the DAVIS+VIDEVO dataset~\cite{lai2018learning}, random crop and rotation are used as standard data augmentation operations, while for the FaceForensics~\cite{rossler2018faceforensics} dataset, only center crop is applied. As for the weights of each loss, we set empirically $\{\lambda_s,\lambda_d,\lambda_r\}: \{\lambda_f,\lambda_p,\lambda_l\} = 10:1 $.

\subsection{Comparison with Existing Methods}

\paragraph{Result quality.}
We compare our method with~\cite{wang2018video}, the only deep video inpainting solution to our best knowledge, both quantitatively and qualitatively, as shown in 
Table~\ref{tab:accuracy} and Figures~\ref{fig:results} and~\ref{fig:teaser} respectively. 
We also include more video results in our accompanying supplemental materials.
Table~\ref{tab:accuracy} compares our model with~\cite{wang2018video} in terms of the average $l_1$ difference on all validation video frames. 
It can be found that our statistic data are better among three mask types and two datasets. It also reveals that the quality gap between~\cite{wang2018video} and us is larger on the natural dataset DAVIS+VIDEVO with larger motion, where the average $l_1$ loss is 40.87 and 13.89, in comparison with the values 8.2 and 6.1 for the FaceForensics dataset~\cite{rossler2018faceforensics}. This phenomenon potentially results from the fact that our model is designed to work better on large motion videos due to ConvLSTM and optical flow as involved, while 3DCNN in~\cite{wang2018video}, has very limited capability to capture motion. Furthermore, the quality gap on random walker masks between~\cite{wang2018video} and ours is larger than those on the two rectangle masks, which demonstrates that our model can better deal with masks with various shapes. The last but not the least, it is worth noting that with one single Titan Xp GPU, our model can deal with videos of arbitrary length, and of frame size as high as 256p in real time\footnote{For fair comparison, we limit the test videos to be of 32 frames and 128p for our method in the experiments.}, in comparison with~\cite{wang2018video} which can handle videos with fixed length of 32 frames and size of 128p only.

As for qualitative results, it is also obvious that our results outperform~\cite{wang2018video} as Figure~\ref{fig:results} shows. The difference between our results and~\cite{wang2018video} on FaceForensics mainly lies in the sharpness of the results. With the 3DCNN as a temporal inference approach, it seems pixels from multiple frames are blended without being aware of the location movement.
As for the results of DAVIS+VIDEVO,~\cite{wang2018video} barely fills in reasonable patches, making the border easily noticeable in contrary with ours. 

\paragraph{Computational efficiency.}

We also include several other metrics such as the time and memory efficiency to compare~\cite{wang2018video}. We calculate the average inference time per video and the number of trainable parameters in both models. 
As seen from Table~\ref{tab:efficiency}, without 3DCNN module involved, our model achieves better time and memory efficiency compared with~\cite{wang2018video}, i.e. our inference time and number of parameters is only around $2/3$ and $1/5$ of CombCN. {\color{black}{Also thanks to the fully convolutional architecture of the model, our approach can accept videos of larger frame size and arbitrary length.}}


\subsection{Ablation Studies}

\input{tab-b1-b2-lstm-pc-ours.tex}
We further conduct the following ablation study to discover to influence of each part in our network towards the final results.
\paragraph{Ours vs. PartialConv.}\label{sec:vspartial}
We first compare our results $\{O_t\}$ with those produced by PartialConv~\cite{liu2018image} frame-wisely, i.e. $\{P_t\}$. The statistic data is listed in the Columns "PartialConv" and "Ours" in Table~\ref{tab:b1-b2-lstm-pc-ours} and some samples are shown in Figure~\ref{fig:ablation} accordingly. As seen, our approach achieves higher quality compared with PartialConv-only method, especially for the cases of random rectangle and random walker masks. This is due to the fact that by using ConvLSTM, our method can model the temporal coherence so that information from adjacent frames could be learnt. Even though a region is damaged in the current frame, similar patches could be found in its adjacent frames due to random masks moving away. Meanwhile, our method better improves the temporal coherence and reconstructed details in DAVIS+VIDEVO than FaceForensics dataset, which lacks enough motion and diversity. This phenomenon demonstrates the effectiveness of ConvLSTM module, which enables our method to preserve the temporal consistency across frames and to be specialized in inpainting videos with more motion.

\paragraph{Ours vs. ConvLSTM.}
We further validate the necessity of inpainting all frames by frame-wise inpainting module $\mathcal{H}_s$. Specifically, we instead inpaint the first and last frame only in a video by PartialConv, producing $P_t = \mathcal{H}_s(I_t), t \in \{1, T\}$ and keep the rest frames untouched. We then feed the sequence $\{P_1,I_2,I_3,...,I_{T-1},P_T\}$ to the ConvLSTM module for training and validation, and show the results in Table~\ref{tab:b1-b2-lstm-pc-ours} and Figure~\ref{fig:ablation}. As seen, without the PartialConv module, ConvLSTM produces low-quality results compared with ours, especially for fixed rectangle masks. The reason behind is that, fixed rectangle masks make the missing regions constant across frames, in which case ConvLSTM has very limited information from adjacent frames to inpaint the current frame. 

\paragraph{}From the two comparisons as above, we see that our method generally outperforms each single module of the entire model as proposed. With PartialConv only, the produced results are of satisfied quality for a single frame, yet look incompatible when being played, lacking of stable temporal coherence. On the other hand, with ConvLSTM involved only, temporal information from the adjacent frames could be recovered, but details are hardly reconstructed, causing the filled region inconsistent with the content of the present frame. This is due to that the strength of ConvLSTM is on temporal information recovery but it lacks enough capability of modeling spatial details. Our approach naturally combine the advantages of the two modules, so that benefits from the strengths of both of them.

\paragraph{Disabling the flow blending network $\mathcal{H}_f$.}\label{par:one-flow}
To verify the effectiveness of the flow blending network $\mathcal{H}_f$ as proposed, we alternatively disable $\mathcal{H}_f$ and set $F_{t-1,t}$ to $F^P_{t-1,t}$ or $F^I_{t-1,t}$ directly. For these two cases, we additionally list their $l_1$ losses in Table~\ref{tab:b1-b2-lstm-pc-ours} to compare with ours. We see that without the flow blending network, the performance drops on almost all datasets and mask types. With the flow $F^I_{t-1,t}$ generated by the flow completion network $\mathcal{H}_c$, the misleading information in $F^P_{t-1,t}$ is corrected, which enables our approach to produce more robust inpainted frames. We also accompany the inpainted videos by using $F^I_{t-1,t}$ and $F^P_{t-1,t}$ only in our supplementary materials for better presentation, where more noticeable flickering artifacts exist.

%% file: results_fig.tex
\begin{figure*}[t]
  \begin{center}
  \includegraphics[width=0.98\linewidth]{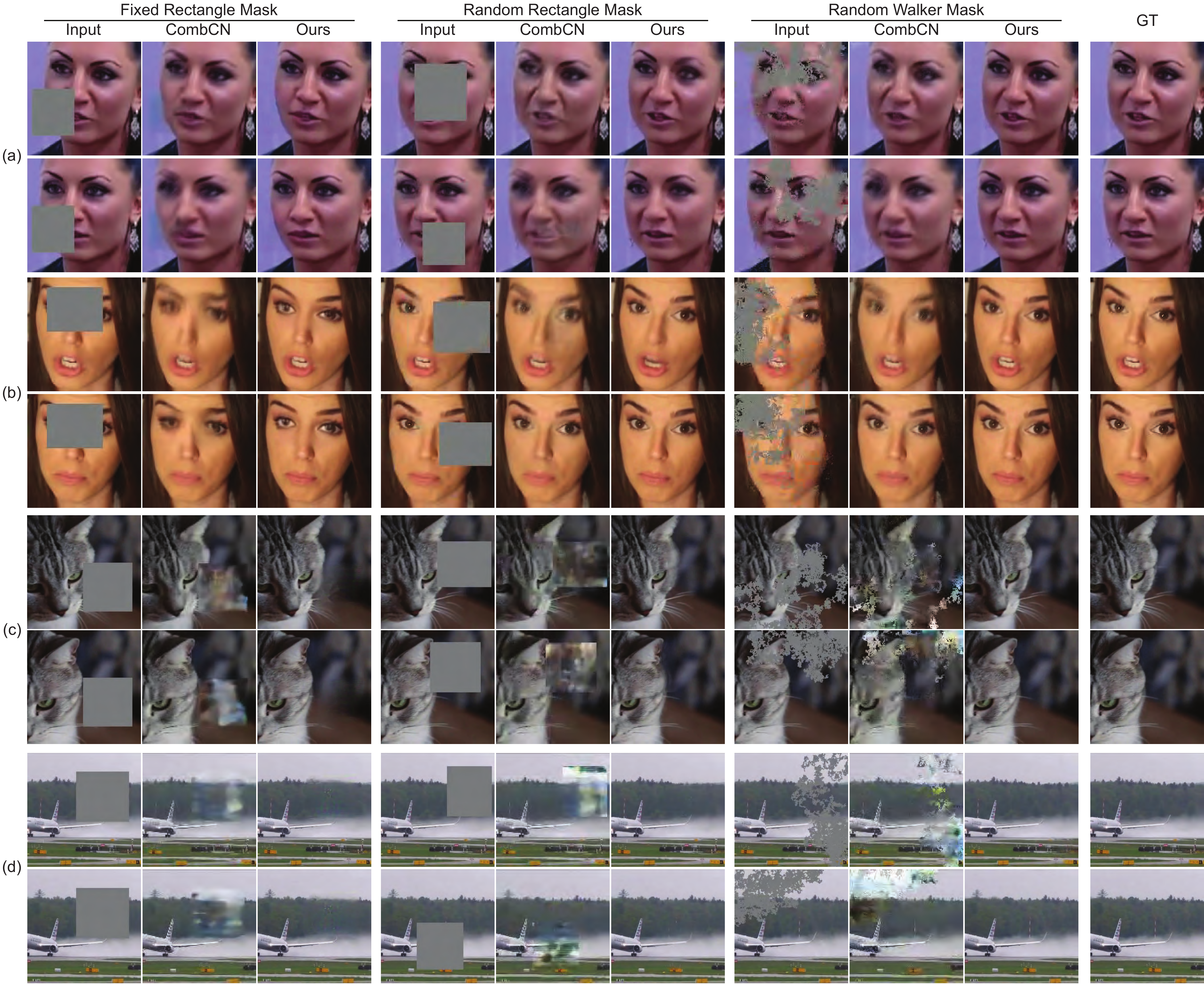}
  \end{center}
  \caption{Results: The first 2 samples are from FaceForensics~\cite{rossler2018faceforensics} and the rest are from DAVIS+VIDEVO~\cite{lai2018learning}. The rows shows different frames in a video, and the column shows the results comparison under different masks. Checking the columns marked as "CombCN" and "Ours", it can be found that our results outperform CombCN in all mask and dataset combinations, especially on the complex natuaral dataset DAVIS+VIDEVO~\cite{lai2018learning} and random masks. }\label{fig:results}
\end{figure*}

%% file: tab-accuracy.tex
\begin{table}[t]
\begin{center}
\resizebox{0.98\linewidth}{!}{
    \begin{tabular}{l
    >{\centering\arraybackslash}p{1.5cm}
    >{\centering\arraybackslash}p{1.5cm}
    >{\centering\arraybackslash}p{1.5cm}
    >{\centering\arraybackslash}p{1.5cm}}
    \toprule
    \multirow{2}[2]{*}{} & \multicolumn{2}{c}{FaceForensics} & \multicolumn{2}{c}{DAVIS+VIDEVO} \\
          & \multicolumn{1}{c}{CombCN~\cite{wang2018video}} & \multicolumn{1}{c}{Ours} & \multicolumn{1}{c}{CombCN~\cite{wang2018video}} & \multicolumn{1}{c}{Ours} \\
    \midrule
    Fixed Rectangles & 13.68 & \textbf{12.03} & 37.63 & \textbf{23.38} \\
    \midrule
    Random Rectangles & 5.84  & \textbf{4.01} & 42.41 & \textbf{12.27} \\
    \midrule
    Random Walker~\cite{sundaram2010dense} & 5.07  & \textbf{2.25} & 42.57 & \textbf{6.01} \\
    \midrule
    Average & 8.20  & \textbf{6.10} & 40.87 & \textbf{13.89} \\
    \bottomrule
    \end{tabular}}
\end{center}
    \caption{$l_1$ loss on dataset FaceForensics and DAVIS+VIDEVO.}\label{tab:accuracy}%
\end{table}%

%% file: tab-efficiency.tex
%

\begin{table}[t]
\begin{center}
\resizebox{0.98\linewidth}{!}{
    \begin{tabular}{p{4.1cm}>{\centering\arraybackslash}p{2.1cm}>{\centering\arraybackslash}p{3.5cm}}
    \toprule
    Model & CombCN~\cite{wang2018video} & Ours \\
    \midrule
    Inference Time (ms/frame) & 36.06 & \textbf{23.31} (35.35\% less) \\
    \midrule
    No. of Parameters & 15,489,539 & \textbf{2,947,011} (80.97\% less) \\
    \midrule
    Input Length & fixed & arbitrary \\
    \midrule
    Input Resolution & 128   & arbitrary* \\
    \bottomrule
    \end{tabular}}
\end{center}
\caption{Comparison in time and memory efficiency between CombCN and our method. * means our model is fully convolutional and not restricted to the input size.}\label{tab:efficiency}
\end{table}%

%% file: ours-vs-lstm-pc.tex
\begin{figure}[t]
\begin{center}
  \includegraphics[width=0.98\linewidth]{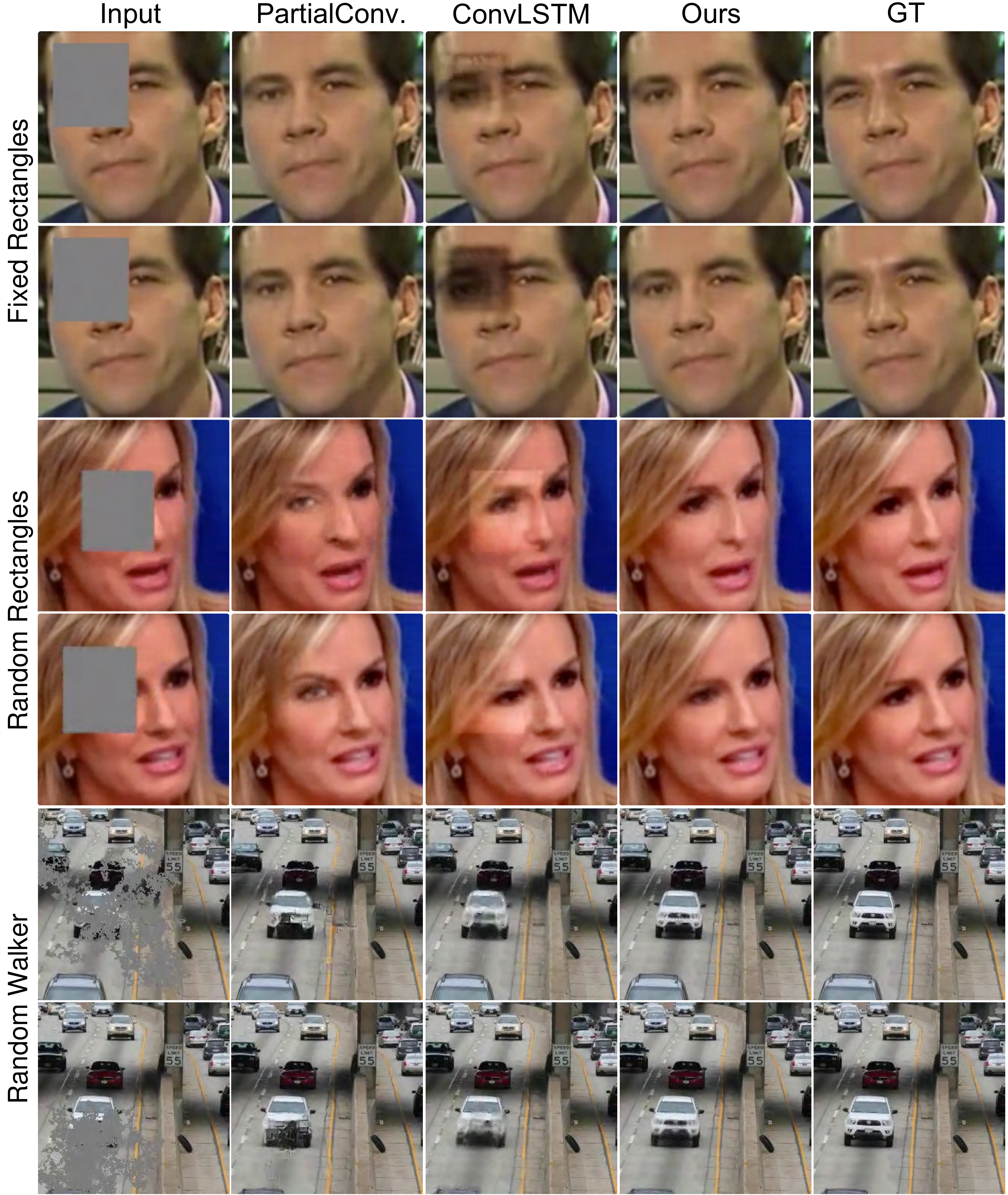}
  \end{center}
  \caption{Ablation study results. We list 3 samples, each of which contains 2 frames from top to bottom rows, under the conditions of fixed rectangles, random rectangles and random walker masks, respectively. In the 2nd and 3rd columns, we illustrate the results by PartialConv-only and ConvLSTM-only modules. Compared with ours, these two models produce low quality results in terms of time consistency and spatial details. }\label{fig:ablation}

\end{figure}

%% file: tab-b1-b2-lstm-pc-ours.tex
\begin{table*}[t]
\begin{center}
\resizebox{0.98\linewidth}{!}{
    \begin{tabular}{l
    >{\centering\arraybackslash}p{1.6cm}
    >{\centering\arraybackslash}p{1.6cm}
    >{\centering\arraybackslash}p{1.3cm}
    >{\centering\arraybackslash}p{1.3cm}
    >{\centering\arraybackslash}p{1.2cm}
    >{\centering\arraybackslash}p{1mm}
    >{\centering\arraybackslash}p{1.6cm}
    >{\centering\arraybackslash}p{1.6cm}
    >{\centering\arraybackslash}p{1.3cm}
    >{\centering\arraybackslash}p{1.3cm}
    >{\centering\arraybackslash}p{1.2cm}}
    \toprule
    \multirow{2}[4]{*}{} & \multicolumn{5}{c}{FaceForensics~\cite{rossler2018faceforensics}}     &       & \multicolumn{5}{c}{DAVIS+VIDEVO~\cite{lai2018learning}} \\
\cmidrule{2-6}\cmidrule{8-12}          & PartialConv & ConvLSTM  & $F_{t-1,t}^P$ & $F_{t-1,t}^I$ & Ours &       & PartialConv & ConvLSTM  & $F_{t-1,t}^P$ & $F_{t-1,t}^I$ & Ours \\
    \midrule
    Fixed Rectangles & 11.59  & 38.08  & \textbf{11.54 } & 12.22  & 12.03  &       & 24.74  & 53.10  & 24.94  & 27.30  & \textbf{23.38 } \\
    \midrule
    Random Rectangles & 11.28  & 22.68  & 4.13  & 4.73  & \textbf{4.01 } &       & 24.99  & 17.80  & 12.80  & 13.07  & \textbf{12.27 } \\
    \midrule
    Random Walker & 3.31  & 3.75  & 2.35  & 2.75  & \textbf{2.25 } &       & 10.67  & 10.00  & 7.74  & 7.39  & \textbf{6.01 } \\
    \bottomrule
    \end{tabular}
}
\end{center}
\caption{$l_1$ losses of results in ablation study for datasets FaceForensics~\cite{rossler2018faceforensics} and DAVIS+VIDEVO~\cite{lai2018learning}. The ablation study include only using PartialConv, ConvLSTM, flow $F^P_{t-1,t}$ or $F^I_{t-1,t}$.}\vspace{-1mm}\label{tab:b1-b2-lstm-pc-ours}%
\end{table*}%

%% file: conclu.tex
\section{Conclusion}\label{sec:conclu}

We have presented a new video inpainting framework based on ConvLSTM and robust optical flow generation. Our framework can produce inpainted video frames with spatial details and temporal coherence. Unlike the previous volume based solutions~\cite{wang2018video,wang2014video,wang2017video}, our method does not restrict the video length and frame size, being able to run in real-time streamingly, and can deal with large motions. These advantages lie in the strong capability of ConvLSTM in modeling spatial-temporal information simultaneously, and the rich motion information conveyed in the optical flow. To generate an accurate optical flow from frames with holes, we propose a robust flow generation module which can be fed with two sources of flows. A flow blending network is also proposed to learn the mechanism to fuse the two flows, producing results with less error.
We further introduce three different masks and two datasets to thoroughly test our method. Experimental results demonstrate our superior performance in comparison with the state-of-the-art in all of the scenarios as above. Ablation studies are also conducted to reveal the effectiveness of different parts in our framework.

%% file: main.bbl
\begin{thebibliography}{10}\itemsep=-1pt

\bibitem{barnes2009patchmatch}
C.~Barnes, E.~Shechtman, A.~Finkelstein, and D.~B. Goldman.
\newblock Patchmatch: A randomized correspondence algorithm for structural
  image editing.
\newblock In {\em ACM Transactions on Graphics (ToG)}, volume~28, page~24. ACM,
  2009.

\bibitem{barnes2015patchtable}
C.~Barnes, F.-L. Zhang, L.~Lou, X.~Wu, and S.-M. Hu.
\newblock Patchtable: Efficient patch queries for large datasets and
  applications.
\newblock {\em ACM Transactions on Graphics (TOG)}, 34(4):97, 2015.

\bibitem{das2018deep}
S.~Das, M.~Koperski, F.~Bremond, and G.~Francesca.
\newblock Deep-temporal lstm for daily living action recognition.
\newblock {\em arXiv preprint arXiv:1802.00421}, 2018.

\bibitem{efros1999texture}
A.~A. Efros and T.~K. Leung.
\newblock Texture synthesis by non-parametric sampling.
\newblock In {\em Proceedings of the seventh IEEE international conference on
  computer vision}, volume~2, pages 1033--1038. IEEE, 1999.

\bibitem{gao2017video}
L.~Gao, Z.~Guo, H.~Zhang, X.~Xu, and H.~T. Shen.
\newblock Video captioning with attention-based lstm and semantic consistency.
\newblock {\em IEEE Transactions on Multimedia}, 19(9):2045--2055, 2017.

\bibitem{gers1999learning}
F.~A. Gers, J.~Schmidhuber, and F.~Cummins.
\newblock Learning to forget: Continual prediction with lstm.
\newblock 1999.

\bibitem{he2016deep}
K.~He, X.~Zhang, S.~Ren, and J.~Sun.
\newblock Deep residual learning for image recognition.
\newblock In {\em Proceedings of the IEEE conference on computer vision and
  pattern recognition}, pages 770--778, 2016.

\bibitem{hochreiter1997long}
S.~Hochreiter and J.~Schmidhuber.
\newblock Long short-term memory.
\newblock {\em Neural computation}, 9(8):1735--1780, 1997.

\bibitem{IizukaSIGGRAPH2017}
S.~Iizuka, E.~Simo-Serra, and H.~Ishikawa.
\newblock {Globally and Locally Consistent Image Completion}.
\newblock {\em ACM Transactions on Graphics (Proc. of SIGGRAPH 2017)},
  36(4):107:1--107:14, 2017.

\bibitem{ilg2017flownet}
E.~Ilg, N.~Mayer, T.~Saikia, M.~Keuper, A.~Dosovitskiy, and T.~Brox.
\newblock Flownet 2.0: Evolution of optical flow estimation with deep networks.
\newblock In {\em Proceedings of the IEEE Conference on Computer Vision and
  Pattern Recognition}, pages 2462--2470, 2017.

\bibitem{jia2005video}
Y.-T. Jia, S.-M. Hu, and R.~R. Martin.
\newblock Video completion using tracking and fragment merging.
\newblock {\em The Visual Computer}, 21(8-10):601--610, 2005.

\bibitem{johnson2016perceptual}
J.~Johnson, A.~Alahi, and L.~Fei-Fei.
\newblock Perceptual losses for real-time style transfer and super-resolution.
\newblock In {\em European conference on computer vision}, pages 694--711.
  Springer, 2016.

\bibitem{kingma2014adam}
D.~P. Kingma and J.~Ba.
\newblock Adam: A method for stochastic optimization.
\newblock {\em arXiv preprint arXiv:1412.6980}, 2014.

\bibitem{kwatra2005texture}
V.~Kwatra, I.~Essa, A.~Bobick, and N.~Kwatra.
\newblock Texture optimization for example-based synthesis.
\newblock In {\em ACM Transactions on Graphics (ToG)}, volume~24, pages
  795--802. ACM, 2005.

\bibitem{lai2018learning}
W.-S. Lai, J.-B. Huang, O.~Wang, E.~Shechtman, E.~Yumer, and M.-H. Yang.
\newblock Learning blind video temporal consistency.
\newblock In {\em Proceedings of the European Conference on Computer Vision
  (ECCV)}, pages 170--185, 2018.

\bibitem{li2018videolstm}
Z.~Li, K.~Gavrilyuk, E.~Gavves, M.~Jain, and C.~G. Snoek.
\newblock Videolstm convolves, attends and flows for action recognition.
\newblock {\em Computer Vision and Image Understanding}, 166:41--50, 2018.

\bibitem{liu2018image}
G.~Liu, F.~A. Reda, K.~J. Shih, T.-C. Wang, A.~Tao, and B.~Catanzaro.
\newblock Image inpainting for irregular holes using partial convolutions.
\newblock In {\em Proceedings of the European Conference on Computer Vision
  (ECCV)}, pages 85--100, 2018.

\bibitem{liu2016spatio}
J.~Liu, A.~Shahroudy, D.~Xu, and G.~Wang.
\newblock Spatio-temporal lstm with trust gates for 3d human action
  recognition.
\newblock In {\em European Conference on Computer Vision}, pages 816--833.
  Springer, 2016.

\bibitem{meng2018mganet}
X.~Meng, X.~Deng, S.~Zhu, S.~Liu, C.~Wang, C.~Chen, and B.~Zeng.
\newblock Mganet: A robust model for quality enhancement of compressed video.
\newblock {\em arXiv preprint arXiv:1811.09150}, 2018.

\bibitem{nilsson2018semantic}
D.~Nilsson and C.~Sminchisescu.
\newblock Semantic video segmentation by gated recurrent flow propagation.
\newblock In {\em Proceedings of the IEEE Conference on Computer Vision and
  Pattern Recognition}, pages 6819--6828, 2018.

\bibitem{pathak2016context}
D.~Pathak, P.~Krahenbuhl, J.~Donahue, T.~Darrell, and A.~A. Efros.
\newblock Context encoders: Feature learning by inpainting.
\newblock In {\em Proceedings of the IEEE Conference on Computer Vision and
  Pattern Recognition}, pages 2536--2544, 2016.

\bibitem{ren2016look}
J.~Ren, Y.~Hu, Y.-W. Tai, C.~Wang, L.~Xu, W.~Sun, and Q.~Yan.
\newblock Look, listen and learn—a multimodal lstm for speaker
  identification.
\newblock In {\em Thirtieth AAAI Conference on Artificial Intelligence}, 2016.

\bibitem{ronneberger2015u}
O.~Ronneberger, P.~Fischer, and T.~Brox.
\newblock U-net: Convolutional networks for biomedical image segmentation.
\newblock In {\em International Conference on Medical image computing and
  computer-assisted intervention}, pages 234--241. Springer, 2015.

\bibitem{rossler2018faceforensics}
A.~R{\"o}ssler, D.~Cozzolino, L.~Verdoliva, C.~Riess, J.~Thies, and
  M.~Nie{\ss}ner.
\newblock Faceforensics: A large-scale video dataset for forgery detection in
  human faces.
\newblock {\em arXiv preprint arXiv:1803.09179}, 2018.

\bibitem{rumelhart1988learning}
D.~E. Rumelhart, G.~E. Hinton, R.~J. Williams, et~al.
\newblock Learning representations by back-propagating errors.
\newblock {\em Cognitive modeling}, 5(3):1, 1988.

\bibitem{simonyan2014very}
K.~Simonyan and A.~Zisserman.
\newblock Very deep convolutional networks for large-scale image recognition.
\newblock {\em arXiv preprint arXiv:1409.1556}, 2014.

\bibitem{sundaram2010dense}
N.~Sundaram, T.~Brox, and K.~Keutzer.
\newblock Dense point trajectories by gpu-accelerated large displacement
  optical flow.
\newblock In {\em European conference on computer vision}, pages 438--451.
  Springer, 2010.

\bibitem{terwilliger2019recurrent}
A.~Terwilliger, G.~Brazil, and X.~Liu.
\newblock Recurrent flow-guided semantic forecasting.
\newblock In {\em 2019 IEEE Winter Conference on Applications of Computer
  Vision (WACV)}, pages 1703--1712. IEEE, 2019.

\bibitem{venkatesh2009efficient}
M.~V. Venkatesh, S.-c.~S. Cheung, and J.~Zhao.
\newblock Efficient object-based video inpainting.
\newblock {\em Pattern Recognition Letters}, 30(2):168--179, 2009.

\bibitem{wang2014video}
C.~Wang, Y.~Guo, J.~Zhu, L.~Wang, and W.~Wang.
\newblock Video object co-segmentation via subspace clustering and quadratic
  pseudo-boolean optimization in an mrf framework.
\newblock {\em IEEE Transactions on Multimedia}, 16(4):903--916, 2014.

\bibitem{wang2018video}
C.~Wang, H.~Huang, X.~Han, and J.~Wang.
\newblock Video inpainting by jointly learning temporal structure and spatial
  details.
\newblock {\em arXiv preprint arXiv:1806.08482}, 2018.

\bibitem{wang2017video}
C.~Wang, J.~Zhu, Y.~Guo, and W.~Wang.
\newblock Video vectorization via tetrahedral remeshing.
\newblock {\em IEEE Transactions on Image Processing}, 26(4):1833--1844, 2017.

\bibitem{wang2019gif2video}
Y.~Wang, H.~Huang, C.~Wang, T.~He, J.~Wang, and M.~Hoai.
\newblock Gif2video: Color dequantization and temporal interpolation of gif
  images.
\newblock {\em arXiv preprint arXiv:1901.02840}, 2019.

\bibitem{wexler2004space}
Y.~Wexler, E.~Shechtman, and M.~Irani.
\newblock Space-time video completion.
\newblock In {\em Proceedings of the 2004 IEEE Computer Society Conference on
  Computer Vision and Pattern Recognition, 2004. CVPR 2004.}, volume~1, pages
  I--I. IEEE, 2004.

\bibitem{wexler2007space}
Y.~Wexler, E.~Shechtman, and M.~Irani.
\newblock Space-time completion of video.
\newblock {\em IEEE Transactions on pattern analysis and machine intelligence},
  29(3), 2007.

\bibitem{xie2012image}
J.~Xie, L.~Xu, and E.~Chen.
\newblock Image denoising and inpainting with deep neural networks.
\newblock In {\em Advances in neural information processing systems}, pages
  341--349, 2012.

\bibitem{xingjian2015convolutional}
S.~Xingjian, Z.~Chen, H.~Wang, D.-Y. Yeung, W.-K. Wong, and W.-c. Woo.
\newblock Convolutional lstm network: A machine learning approach for
  precipitation nowcasting.
\newblock In {\em Advances in neural information processing systems}, pages
  802--810, 2015.

\bibitem{xu2015show}
K.~Xu, J.~Ba, R.~Kiros, K.~Cho, A.~Courville, R.~Salakhudinov, R.~Zemel, and
  Y.~Bengio.
\newblock Show, attend and tell: Neural image caption generation with visual
  attention.
\newblock In {\em International conference on machine learning}, pages
  2048--2057, 2015.

\bibitem{yamaguchi2018high}
S.~Yamaguchi, S.~Saito, K.~Nagano, Y.~Zhao, W.~Chen, K.~Olszewski,
  S.~Morishima, and H.~Li.
\newblock High-fidelity facial reflectance and geometry inference from an
  unconstrained image.
\newblock {\em ACM Transactions on Graphics (TOG)}, 37(4):162, 2018.

\bibitem{yang2017high}
C.~Yang, X.~Lu, Z.~Lin, E.~Shechtman, O.~Wang, and H.~Li.
\newblock High-resolution image inpainting using multi-scale neural patch
  synthesis.
\newblock In {\em The IEEE Conference on Computer Vision and Pattern
  Recognition (CVPR)}, volume~1, page~3, 2017.

\bibitem{zhao2018identity}
Y.~Zhao, W.~Chen, J.~Xing, X.~Li, Z.~Bessinger, F.~Liu, W.~Zuo, and R.~Yang.
\newblock Identity preserving face completion for large ocular region
  occlusion.
\newblock {\em British Machine Vision Conference (BMVC)}, 2018.

\bibitem{zhou2017spatio}
K.~Zhou, Y.~Zhu, and Y.~Zhao.
\newblock A spatio-temporal deep architecture for surveillance event detection
  based on convlstm.
\newblock In {\em 2017 IEEE Visual Communications and Image Processing (VCIP)},
  pages 1--4. IEEE, 2017.

\end{thebibliography}
